\documentclass[twoside]{article}

\usepackage[accepted]{aistats2026}

\usepackage{multirow}
\usepackage{float}
\usepackage{epsfig}
\usepackage{epstopdf}
\usepackage{cite}
\usepackage{subcaption}
\usepackage{rotating}
\usepackage{dblfloatfix}
\usepackage{textcomp}
\usepackage{array}
\usepackage{comment}
\usepackage{amsmath}
\usepackage{xcolor}
\definecolor{newcolor}{rgb}{.8,.349,.1}
\usepackage{url}
\usepackage{hyperref}

\graphicspath{{./Images/}}

\begin{document}

\twocolumn[

\aistatstitle{The Unseen Adversaries: Robust and Generalized Defense Against Adversarial Patches}

\aistatsauthor{Vishesh Kumar, Akshay Agarwal}

\aistatsaddress{Trustworthy BiometraVision Lab, \\ Indian Institute of Science Education and Research Bhopal, India}]

\begin{abstract}
  The vulnerabilities of deep neural networks against singularities have raised serious concerns regarding their deployment in the physical world. One of the most prominent and impactful physical-world adversarial perturbations is the attachment of patches to clean images, known as an adversarial patch attack. Similarly, natural noises such as Gaussian and Salt\&Pepper are highly prevalent in the real world. The current research need arises from the above vulnerabilities and the lack of efforts to tackle these two singularities independently and, especially, in combination. In this research, we have, for the first time, combined these two prominent singularities and proposed a novel dataset. Using this dataset, we have conducted a benchmark study of singularity data-point detection using features from several convolutional neural networks. For classification, rather than the popular neural network-based parameter tuning, we have used traditional yet effective machine learning classifiers. The extensive experiments across various in- and out-of-distribution (OOD) singularities reveal several interesting findings about the effectiveness of classifiers and show that it is hard to defend against adversaries when they are treated independently, and inefficient classifiers are selected. 
\end{abstract}

\begin{table*}[!ht]
    \centering
    \caption{ Review of recent papers on adversarial patch attack generation and detection algorithms along with defense against natural noises.}
    \label{survey}
    \setkeys{Gin}{keepaspectratio}
\resizebox*{0.99\textwidth}{0.99\textheight} {
    \begin{tabular}{|l|l|l|}
    \hline
        Adversary  & Authors & Description \\ \hline
        \multirow{5}{*}{\begin{tabular}[c]{@{}l@{}}Patch \\ Generation \end{tabular}} & \cite{sun2023d}  &  Diversified Universal Adversarial Patch Generation Method (D-UAP). \\ \cline{2-3}
        ~ &\cite{he2023generating} & Imperceptible adversarial patch - Vulnerable target category + target attack. \\ \cline{2-3}
        ~ & \cite{rasol2023adaptive} & Adaptive adversarial patch-generating network, AAPGNet. \\ \cline{2-3}
        ~ & \cite{zhou2023downstream} & Adversarial patch for a set of natural images using AdvEncoder. \\ \cline{
        2-3
        }
        ~ & \cite{tang2023adversarial} & Generate adversarial patches against aerial imagery detectors.  \\ \hline \hline
        
        \multirow{6}{*}{\begin{tabular}[c]{@{}l@{}}Patch \\ Detection \end{tabular}} & \cite{ojaswee2023benchmarking} & Fintuned deep neural networks for patch detection. \\ \cline{
        2-3
        }
        ~ & \cite{tarchoun2023jedi} & A differential entropy analysis to detect adversarial patches. \\ \cline{
        2-3
        }
        ~ & \cite{xu2023patchzero} & PatchZero defence, Zeros out patch region. \\ \cline{2-3}
        ~ & \cite{kang2023diffender} & Diffusion-based defense by localization of patch. \\ \cline{2-3}
        ~ & \cite{liu2022segment} & Segment and complete defense to detect and remove the adversarial patch.  \\ \cline{2-3}
        ~ & \cite{xiang2022patchcleanser} & Twice pixel masking to neutralize patch using PatchCleanser. \\ \cline{2-3}
        ~ & \cite{kim2022defending} & APE masking + APE refinement. \\ \hline \hline 
        \multirow{5}{*}{\begin{tabular}[c]{@{}l@{}}Defense \\ Natural \\ Noise \end{tabular}} & \cite{yao2023towards}& Interactive self-supervised denoising with user preferences. \\ \cline{2-3}
        ~ & \cite{wang2022blind2unblind} &  Self-supervised denoising from  single noisy image. \\ \cline{2-3}
        ~ &  \cite{yang2022self} & Improve recognition in low-quality images by using self-feature distillation. \\ \cline{2-3}
        ~ & \cite{zhang2022self}& Jointly improve restoration using dual exposures. \\ \cline{2-3}
        ~ & \cite{zhang2022idr}&  Unsupervised denoising without clean pairs. \\ \hline
    \end{tabular}
    }
\end{table*}

\section{INTRODUCTION}

The sensitivity of automated computer systems, including machine learning algorithms, against adversarial attacks is a serious concern \cite{li2023future}. Among several known adversarial attacks, an adversarial patch attack is one of the stealthiest and most realistic. Physical adversarial patches are small and typically printed as posters or stickers, then placed on target objects in a scene. These patches are also observed to be agnostic to affine transformations, such as translation and rotation, of both target objects and patches. In 2017, \cite{brown2017adversarial} introduced the concept of adversarial patch attacks and demonstrated how they could be used to fool object detectors in the real world. Since then, several advances have been made in developing sophisticated adversarial patches to fool computer vision models. \cite{karmon2018lavan} proposed LaVAN (Localized and visible adversarial noise), which focuses on exploiting the weaknesses of object detectors and creating stealthy patches. Adversarial QR patches \cite{chindaudom2020adversarialqr,chindaudom2022surreptitious} are introduced to make patches less suspicious. \cite{liu2019perceptual} designed PS-GAN (Perceptual Sensitive Generative Adversarial Networks) to improve adversarial patches' visual quality and effectiveness. \cite{gittings2019robust} used deep image priors to create imperceptible perturbations that can still fool object detectors. \cite{zhou2021data} proposed DiAP, a data-independent adversarial patch technique. \cite{lee2019physical} improved DPatch for real-world applications and lighting conditions. \cite{wu2020dpattack} introduced DPAttack (Diffused Patch Attacks) to perturb small image areas effectively, and \cite{huang2021rpattack} extended this with RPAttack (Refined Patch Attack). \cite {thys2019fooling} targeted surveillance cameras, and \cite{den2020adversarial} focused on military asset camouflage. \cite{lu2021scale} introduced Patch-Noobj to scale patches adaptively. \cite{li2019adversarial} developed the Dynamic Adversarial Patch for dynamic scenes. \cite{zolfi2021translucent} introduced the translucent patch for camera lenses, and \cite{wang2021towards} designed the invisibility patch for specific class attacks. \cite{lennon2021patch} introduced mAST (mean Attack Success over Transformations), evaluating patch attack robustness for 3D transformations. \cite{lang2021attention} proposed AGAP using heat maps for patch generation, but faced challenges with feature density in real-world scenarios.

The above review shows that several adversarial patch generation algorithms exist in the literature, and they are effective for almost every computer vision task. However, a lack of patch datasets for benchmarking patch detection significantly limits the development of defense algorithms. Recently, \cite{pintor2023imagenet} and \cite{ojaswee2023benchmarking} have proposed a benchmark adversarial patch dataset to push research towards defending against these effective physical-world singularities of deep networks. Apart from these artificial adversaries, the real world is also affected by several natural noises that predominantly occur in the real world \cite{agarwal2022benchmarking,hendrycks2021natural,pei2019effects}, which the above research ignores. And the combination of these adversarial patches and natural noise singularities further exacerbates neural network degradation, leading to significantly worse performance, as demonstrated in Section \ref{sec: attack_effect}. However, the vulnerabilities of defense algorithms against out-of-distribution samples (e.g., unseen noise or adversarial patches) or the independent handling of singularities result in shallow security algorithms.  

While a limited benchmark patch or noise dataset exists, several researchers have prepared in-house datasets to defend against adversarial patches and natural noise. For example, \cite{gittings2020vax} proposed a training-time defense against patch attacks, introducing VaN (Vaxa-Net), which used a DC-GAN (Deep Convolutional Generative Adversarial Network) \cite{radford2015unsupervised} to generate effective adversarial patches and trained models to defend against them. \cite{radford2015unsupervised} developed an adversarial training technique that improved model robustness against adversarial patches without compromising clean accuracy. Most empirical defenses against adversarial patches rely on adversarial training or saliency map inference, but \cite{cosgrove2020robustness} introduced a distinct approach: CompNets, an interpretable compositional model that inherently resists occlusions and defends against adversarial patches.\cite{huang2021zero} developed PatchVeto, a zero-shot certified defense based on Visual Transformers (ViT) \cite{dosovitskiy2020image}, while \cite{salman2022certified} leveraged ViTs for certified patch attacks. Table \ref{survey} provides a summary of the recent works on adversarial patch generation, detection, and defense against natural noises.

\begin{figure*}[t]
\centering
  \includegraphics[width = 0.9\textwidth]{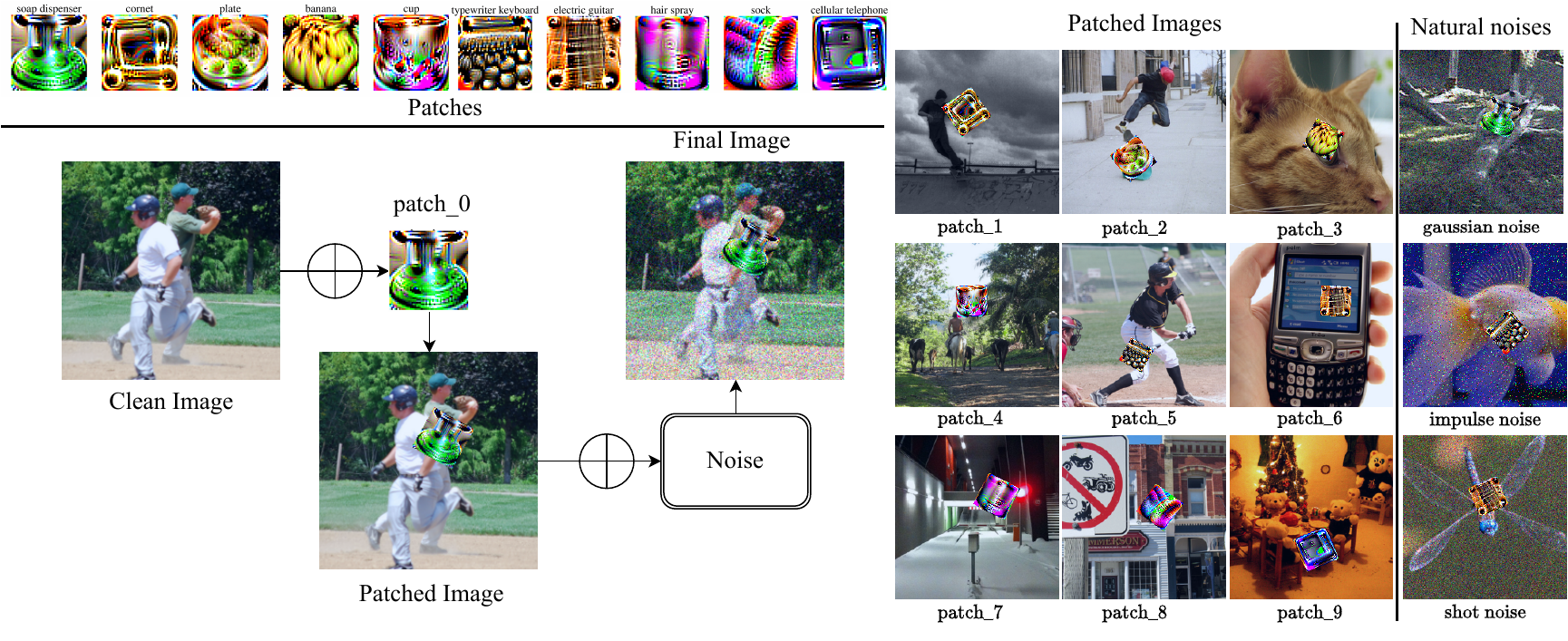}
  \caption{(Left) Overview of the steps involved in the generation of the proposed adversarial patch + natural noise dataset. (Right) Samples images of $10$ patches and patch + $3$ natural noises of the proposed dataset.}
    \label{fig:dataset}
\end{figure*}

As mentioned above, the absence of an adversarial patch benchmark dataset limits the development of a unified defense. Furthermore, the combination of adversarial patches and natural noise remains unexplored. Inspired by the above limitations and the impact of both singularities \cite{11098661, kumar2025unified}, in this research, for the first time we have proposed the singularity dataset, which inherits both adversarial patches and natural noises. We assert that developing such a dataset will advance adversarial patch defense and ensure that the resulting defenses are resilient enough to operate in the noisy, unconstrained physical world. By utilizing the proposed dataset, we have conducted a benchmark study to detect the adversarial patch-based singularity both in the presence and absence of natural noises. In contrast to existing neural network-based defenses, we have plugged in traditional yet effective machine learning classifiers, such as support vector machines, Bayes, and decision forests, for detection. The prime reason is that, firstly, these classifiers are underexplored for adversarial defense, and secondly, they are found to be less susceptible to noisy image alterations than neural networks \cite{9207872,liu2019detection}. In brief, the contributions of this research are: (i) Proposed first-ever patch+noise singularity datasets containing images of multiple variations of physical patches and different natural noises together; and (ii) We comprehensively evaluated different deep neural networks, including the Vision Transformer (ViT) \cite{dosovitskiy2020image} and machine learning classifiers to ensure the development of a unified and resilient defense algorithm.

We observed that these singularities are not limited to classification models (white or black-box) but are task-agnostic; they can even fool object detection models, even when explicitly developed for classification. Therefore, the proposed benchmark study is an important step towards a unified defense and responsible AI deployment.

\begin{table}[t]
    \centering
 \caption{Characteristics of the proposed dataset along with the statistics of images used in training and testing.}
     \label{table_d}
     \setkeys{Gin}{keepaspectratio}
 \resizebox*{0.47\textwidth}{0.47\textheight} {
    \begin{tabular}{|l|l|l|c|c|c|}
    \hline
        Type & Dataset & Adversary & Train & Test & \begin{tabular}[c]{@{}l@{}}Total \end{tabular} \\ \hline
        \multirow{2}{*}{\begin{tabular}[c]{@{}l@{}}Clean \\ (Real)\end{tabular}} & ImageNet & \multirow{2}{*}{--} & 1200 & 800 & 2000 \\ \cline{2-2} \cline{4-6}
        ~ & COCO & ~ & 1200 & 800 & 2000 \\ \hline
        \multirow{6}{*}{Attack} & ImageNet & \multirow{2}{*}{10-Patches} & 12000 & 8000 & 20000 \\ \cline{2-2} \cline{4-6}
        ~ & COCO & ~ & 12000 & 8000 & 20000 \\ \cline{2-6}
         & ImageNet & \multirow{2}{*}{3 - Noise} & 0 & 2400 & 2400 \\ \cline{2-2} \cline{4-6}
        ~ & COCO & ~ & 0 & 2400 & 2400 \\ \cline{2-6}
        & ImageNet & \multirow{2}{*}{\begin{tabular}[c]{@{}l@{}}10-Patch +\\ 3-Noises\end{tabular}} & 0 & 24000 & 24000 \\ \cline{2-2} \cline{4-6}
        ~ & COCO & ~ & 0 & 24000 & 24000 \\ \hline
    \end{tabular}}
    \vspace{-3mm}
\end{table}

\section{PROPOSED ADVERSARIAL DATASET}

The study introduces two novel datasets focusing on physical adversarial patches and natural noises to address the lack of research on developing a unified defense. For that, we have used two popular object recognition datasets, namely ImageNet \cite{deng2009imagenet} and COCO \cite{lin2014microsoft}. The images in these datasets are attacked with $10$ different styles of physical adversarial patches and three types of natural noise (\textit{Gaussian, Shot, and Impulse noise}). To construct adversarial patch images for ImageNet, we randomly selected $2,000$ images from the validation set (not cherry-picked). These images act as a clean subset in the proposed dataset. Subsequently, another set of $2,000$ images from ImageNet is selected, and on top of each image, $10$ selected adversarial patches \cite{pintor2023imagenet} are applied. It generates $20,000$ adversarial patch images along with $2,000$ clean images from the ImageNet dataset. A similar procedure has been applied to the images of the COCO dataset, resulting in $2,000$ clean and $20,000$ adversarial patch images.

To evaluate the robustness of adversarial patch detectors, we have applied natural noise (\textit{Gaussian, shot, and impulse noise}) to the test subset of each dataset. For that, images from clean, individual patches are divided into training and test sets in the ratio $3:2$. For example, there are $2000$ clean images of the ImageNet dataset and $2000$ images of a single patch. When we divide the images into a $3:2$ set, we obtain $800$ test images, and applying three natural noises yields $2,400$ noisy test, clean, and patch images. Since natural noises are inherently present in the physical world \cite{Agarwal_2020_CVPR_Workshops}, they are applied only on the test set to ensure adversarial patch detectors are agnostic to unseen (out-of-distribution (OOD)) singularities. Fig. \ref{fig:dataset} shows the schematic diagram of the proposed patch and patch+noise image generation, and sample images reflecting various patches and noise used in the proposed research. It reflects the challenges in detecting patches due to significant style variations and their tendency to blend into complex image regions.

In total, our dataset contains a diverse set of images, including $4,000$ clean images, and $40,000$ adversarial patched images ($20,000$ from ImageNet and $20,000$ from COCO). Further, $4,800$ test images affected by natural noise and $48,000$ images containing both adversarial patches and natural noise are also present to evaluate the robustness. The characteristics of the proposed dataset are summarized in Table \ref{table_d}.

\subsection{Experimental Setup}

In this research, we first demonstrate the effectiveness of adversarial patches, such as robustness and attack success rate, when combined with natural corruptions, highlighting their impact beyond patch-only scenarios. Then, we have conducted an extensive experimental evaluation across several general settings for the detection of adversarial patches. The first setting is `\textbf{seen patch detection}': in this setting, the patch images from the test set are also used for training. For example, the ImageNet dataset is divided into a training and testing subset where $60$\% of $2000$ images are used for training of clean and patch$_0$ classes and the remaining $40$\% images of clean and patch$_0$ are used for testing. In the `\textbf{unseen patch detection}' setting, patch images used for testing are not seen at the time of training.  For example, $60$\% of $2000$ images are used for training of clean and patch$_0$ classes, and the remaining $40$\% images of clean and patch$_{\neq{0}}$ are used for testing. \textit{In the last setting, we evaluated the robustness of detectors trained on clean and individual patch images (without noises) on the test images (clean and patch classes) perturbed with natural noises}. The training and test splits for each protocol are given in Table \ref{table_d}. Under these settings, the detection performance is analyzed along two main factors: (i) the robustness of deep image encoders (e.g., VGG and ViT) when coupled with traditional classifiers, and (ii) the effectiveness of the training patches in identifying adversarial patches under seen and unseen evaluation scenarios.

Inspired by the preliminary results of \cite{ojaswee2023benchmarking}, we have used two state-of-the-art convolutional neural networks (CNNs), namely VGG16 \cite{simonyan2014very} and Vision Transformer \cite{dosovitskiy2020image}, as a feature extractor. While the authors \cite{ojaswee2023benchmarking} have fine-tuned several CNNs, most are found to be ineffective in generalization settings. Moreover, based on the ineffectiveness of fine-tuning and robustness of traditional classifiers \cite{agarwal2023corruption,agarwal2023parameter,liu2019detection}, we have used several classifiers, including AdaBoost (AB), SGD (Stochastic Gradient Descent), Random Forest (RF), Logistic Regression (LR), and Support Vector Classifier (SVC). These classifiers are trained with the default parameters provided by the Scikit-learn library \cite{sklearn_api}.

\section{RESULT AND ANALYSIS}
As discussed above, we first demonstrate the effectiveness of adversarial patches when combined with natural corruptions, analyzing robust accuracy and attack success rate. These results show that patch+noise combinations lead to a stronger degradation in model robustness than patch-only attacks. Building on these findings, we conduct a comprehensive set of experiments on adversarial patch detection in both seen and unseen settings, evaluated with and without natural noise corruptions.

\begin{table}[t]
\centering
\caption{Mean robust accuracy (\%) of different ImageNet classifiers under adversarial patch attacks and their combination with additive noise corruptions (severity = 2, fixed across all experiments). The lower the value, the better the attack.}
\label{tab:imagenet_patch_noise}
\setkeys{Gin}{keepaspectratio}
\resizebox*{0.49\textwidth}{!}{
{\small
\setlength{\tabcolsep}{5pt}
\renewcommand{\arraystretch}{1.2}
\begin{tabular}{|l|c|c|c|c|c|}
\hline
\multicolumn{1}{|c|}{\textbf{Model}} &
\textbf{Clean Acc.} &
\textbf{Patch Only} &
\textbf{Patch + Gaussian} &
\textbf{Patch + Shot} &
\textbf{Patch + Impulse} \\
\hline \hline
AlexNet        & 64.62 &  8.38 &  2.04 &  2.28 &  \textbf{1.39} \\ \hline
ResNet-18      & 76.08 & 29.78 &  3.40 &  2.86 &  \textbf{1.64} \\ \hline
SqueezeNet-1.0 & 66.62 & 10.45 &  0.41 &  0.48 &  \textbf{0.30} \\ \hline
ResNet-50      & 82.70 & 60.85 & 19.65 & 16.68 &  \textbf{12.49} \\ \hline
MobileNet-V2   & 78.06 & 51.62 &  6.86 &  6.10 &  \textbf{5.30} \\ \hline
VGG-16         & 78.30 & 46.45 &  6.89 &  5.40 &  \textbf{2.72} \\ \hline
ViT-B/16       & 85.10 & 82.68 & 68.23 & 64.88 &  \textbf{62.84} \\ \hline
GoogLeNet      & 76.46 & 47.39 & 15.06 & 14.00 &  \textbf{8.97} \\ \hline
Swin-Tiny      & 85.74 & 83.71 & 66.99 &  \textbf{64.16} & 66.09 \\ \hline
XceptionNet    & 74.94 & 38.88 & 14.27 & 12.95 &  \textbf{9.25} \\
\hline
\end{tabular}
}}
\end{table}

\begin{table}[t]
\centering
\caption{Mean attack success rate (\%) of adversarial patches from \cite{pintor2023imagenet} when applied alone and in combination with additive noise corruptions. The higher the value, the better the attack.}
\label{tab:asr_patch_noise}
\setkeys{Gin}{keepaspectratio}
\resizebox*{0.49\textwidth}{!}{
{\small
\setlength{\tabcolsep}{5pt}
\renewcommand{\arraystretch}{1.2}
\begin{tabular}{|l|c|c|c|c|}
\hline
\multicolumn{1}{|c|}{\textbf{Model}} &
\textbf{Patch Only} &
\textbf{Patch + Gaussian} &
\textbf{Patch + Shot} &
\textbf{Patch + Impulse} \\
\hline \hline
AlexNet        & 26.26 & \textbf{38.03} & 36.89 & 37.91 \\ \hline
ResNet-18      & 46.11 & \textbf{78.31} & 77.50 & 77.20 \\ \hline
SqueezeNet-1.0 & 58.00 & \textbf{77.94} & 77.05 & 75.87 \\ \hline
ResNet-50      &  7.04 & 27.72 & 28.48 & \textbf{30.34} \\ \hline
MobileNet-V2   &  4.49 & 30.73 & \textbf{31.21} & 23.82 \\ \hline
VGG-16         & 11.74 & 39.43 & \textbf{39.60} & 38.88 \\ \hline
ViT-B/16       &  0.71 &  4.14 &  4.95 & \textbf{5.70} \\ \hline
GoogLeNet      & 10.93 & \textbf{31.69} & 30.74 & 30.65 \\ \hline
Swin-Tiny      &  0.67 &  8.93 & \textbf{10.75} &  8.68 \\ \hline
XceptionNet    &  8.33 & 16.14 & \textbf{16.79} & 15.37 \\
\hline
\end{tabular}
}}
\end{table}

\subsection{Attack Effectiveness of Combined Singularities (Patch+Noise)}
\label{sec: attack_effect}

Table \ref{tab:imagenet_patch_noise} shows the mean robust accuracy of 10 ImageNet classifiers under adversarial patches and their combination with natural noise on the ImageNet100 validation set containing 5000 images. The adversarial patches are taken directly from \cite{pintor2023imagenet} and originally optimized against the first three architectures in Table \ref{tab:imagenet_patch_noise} (AlexNet, ResNet-18, and SqueezeNet). Despite this, they transfer well across all models: we observe a clear drop in accuracy under patch-only attacks; for example, ResNet-18 drops from 76.08\% (clean) to 29.78\%, and VGG-16 decreases from 78.30\% to 46.45\%. When adversarial patches are combined with noise (noise + patch), performance degraded further profoundly, with ResNet-18 values reaching 3.40\%, 2.86\%, and 1.64\% accuracy under patch + Gaussian, shot, and impulse noise, respectively, and SqueezeNet falling below 0.5\% across all noise types. Similar trends are observed for other CNNs, such as ResNet-50, which declines from 60.85\% under patch-only attacks to 12.49\% under patch + impulse noise. In contrast, transformer-based architectures such as ViT-B/16 and Swin-Tiny exhibit markedly higher resilience, retaining a large fraction of their patch-only robustness even under combined patch+noise settings. Among the noise types, impulse noise is consistently the most destructive, followed by shot and Gaussian noise, indicating that sparse, high-magnitude perturbations interact more severely with adversarial patches. These trends suggest that adversarial patches act as strong physical-world singularities, and their interaction with natural noise exposes fundamental architectural biases in convolutional models, while global-attention-based models demonstrate comparatively stronger robustness under such compound distribution shifts, although not ready for the real world.

Table~\ref{tab:asr_patch_noise} reports the mean attack success rate of adversarial patches from \cite{pintor2023imagenet} when applied alone and in combination with additive noise corruptions. Consistent with the sharp drop in robust accuracy observed under combined patch+noise settings, the attack success rate increases substantially once applied on noisy images. For example, ResNet-18 exhibits an attack success rate of 46.11\% under patch-only attacks, which rises to 78.31\%, 77.50\%, and 77.20\% when combined with Gaussian, shot, and impulse noise, respectively. A similar pattern is observed for SqueezeNet, where attack success increases from 58.00\% to over 75\% across all patch–noise combinations. Even architectures that appear relatively resistant to patch-only attacks, such as ResNet-50 (7.04\%) and MobileNet-V2 (4.49\%), show a marked increase in vulnerability when noise is added, reaching up to 30.34\% and 31.21\% attack success, respectively. In contrast, transformer-based models demonstrate lower absolute attack success rates; however, the same trend holds, with ViT-B/16 increasing from 0.71\% (patch-only) to 5.70\% under patch + impulse noise, and Swin-Tiny from 0.67\% to 10.75\% under patch + shot noise. These results indicate that additive noise consistently amplifies the effectiveness of adversarial patches across architectures, suggesting that evaluating patch attacks in isolation understates their practical impact under realistic noisy conditions.

\begin{figure}[t]
     \centering
     \begin{subfigure}[b]{0.49\textwidth}
         \centering
         \includegraphics[width=\textwidth]{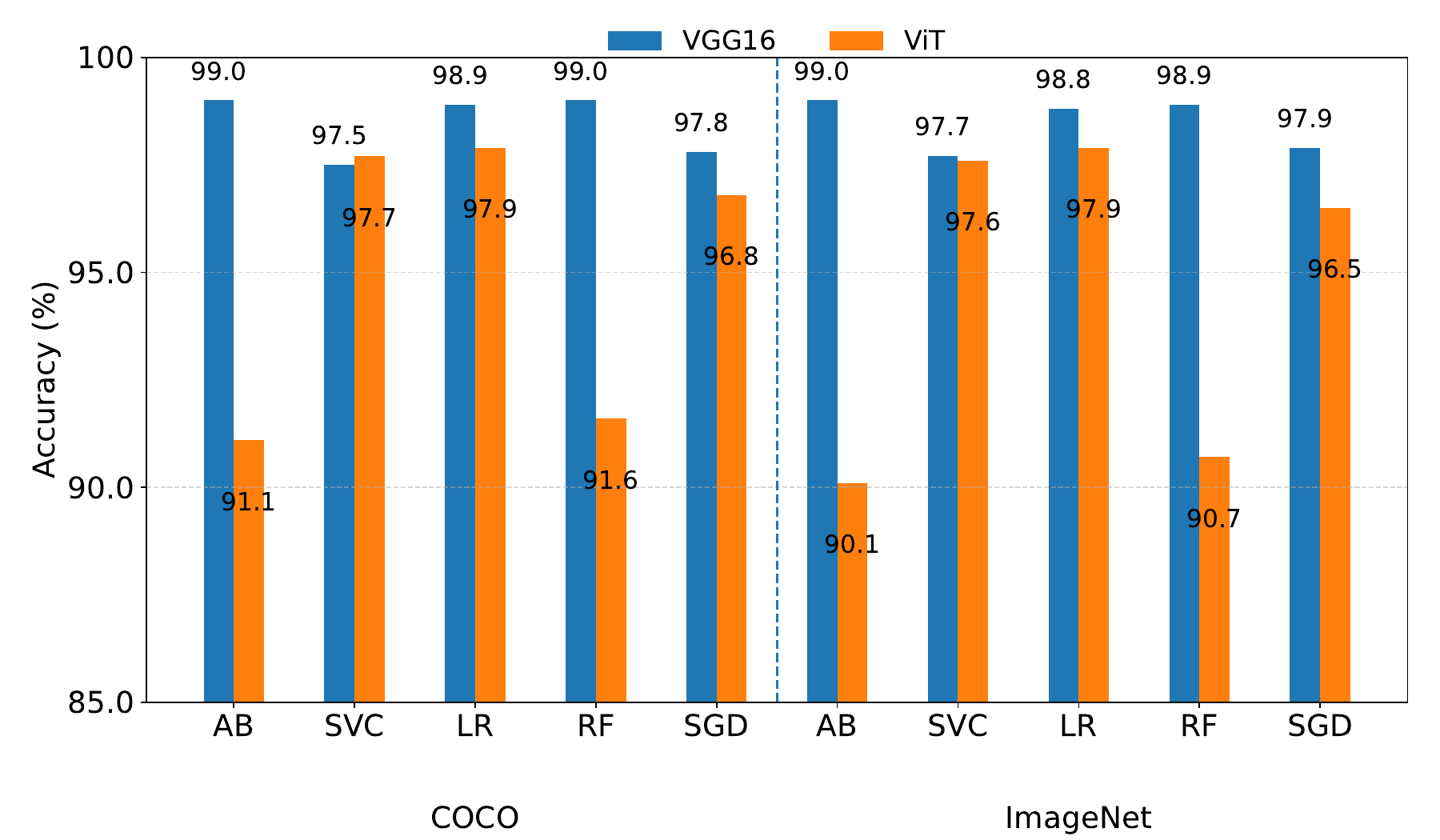}
     \end{subfigure}
     \hfill
     \vspace{-5mm}
        \caption{Average adversarial patch attack detection performance on COCO (left) and ImageNet (right) subset under, seen patches evaluation setting. The results showcase the effectiveness of VGG16 \textbf{when the patches are seen during training and testing}.}
        \label{seenpatch}
\end{figure}

\begin{figure}[t]
     \centering
     \begin{subfigure}[b]{0.49\textwidth}
         \centering
         \includegraphics[width=\textwidth]{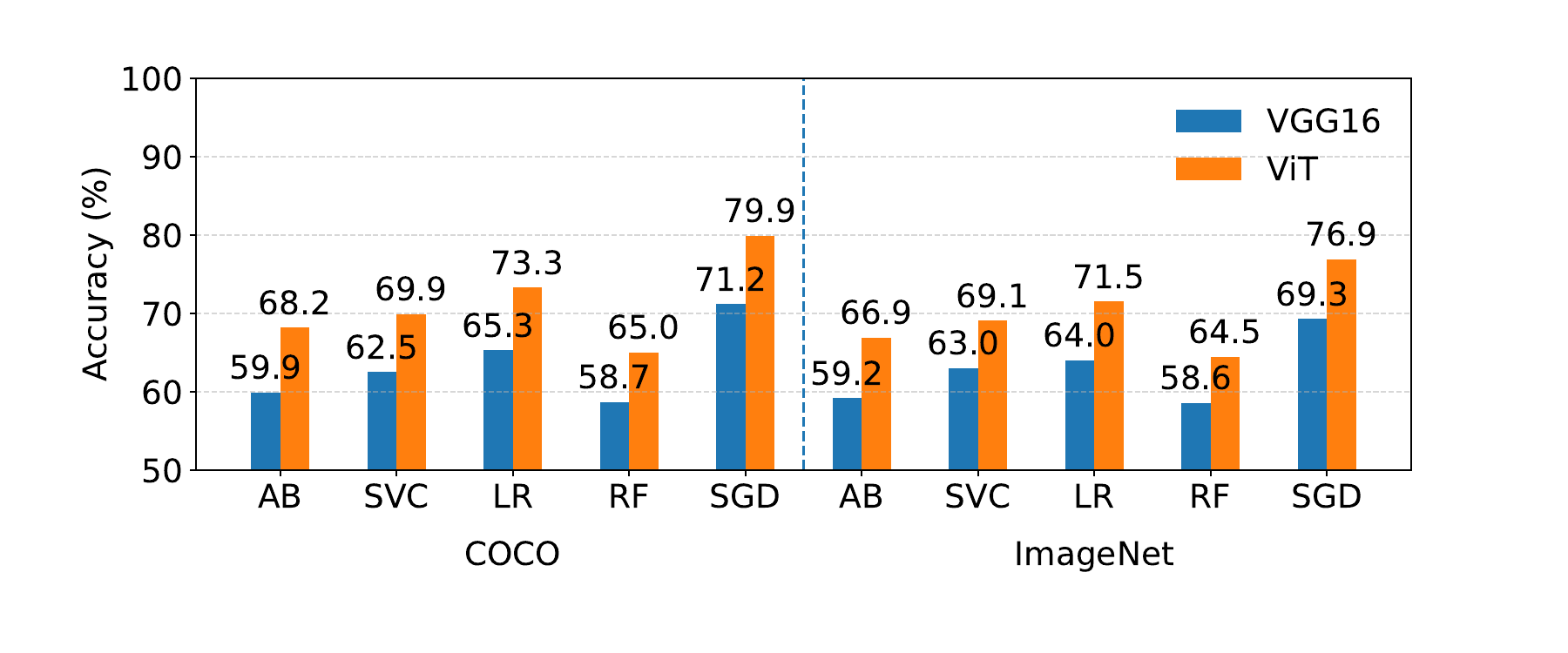}
     \end{subfigure}
     \hfill
     \vspace{-5mm}
        \caption{Average adversarial patch attack detection performance on COCO (left) and ImageNet (right) subset under, unseen patches evaluation setting. The results showcase the generalizability of ViT \textbf{when the patches are unseen during training and testing}.}
        \label{unseenpatch}
\end{figure}

\subsection{Adversarial Patch Detection Analysis}

This section analyzes the effectiveness of adversarial patch detection under both seen and unseen patch settings. We first study detection performance in noise-free conditions and then evaluate the resiliency of the detectors when patches are combined with natural noise perturbations.

\begin{table*}[!ht]
    \centering
\caption{A detailed adversarial patch detection accuracy [0-1] of pure convolution (VGG) and attention (ViT) networks in conjunction with traditional classifiers in \textbf{seen patch detection} setting.}
    \label{coco_ImageNet-seen}
    \setkeys{Gin}{keepaspectratio}
\resizebox*{0.99\textwidth}{0.99\textheight} {
    \begin{tabular}{|l|l|l|l|c|c|c|c|c|c|c|c|c|}
    \hline
        \textbf{Dataset} & \textbf{Models} & \textbf{Classifier} & \textbf{Patch 0} & \textbf{Patch 1} & \textbf{Patch 2} & \textbf{Patch 3} & \textbf{Patch 4} & \textbf{Patch 5} & \textbf{Patch 6} & \textbf{Patch 7} & \textbf{Patch 8} & \textbf{Patch 9} \\ \hline \hline
        
        \multirow{4}{*}{\textbf{\rotatebox{90}{COCO}}} & \multirow{2}{*}{\textbf{VGG16}} & AB & \textbf{0.99} & \textbf{1.00} & \textbf{1.00 }& \textbf{1.00} & \textbf{0.99} & \textbf{0.99} & \textbf{0.99} & \textbf{0.99} & \textbf{0.99} & \textbf{0.99} \\  \cline{3-13}  
        ~ & ~ & SGD & 0.97 & 0.97 & 0.98 & 0.99 & 0.97 & 0.98 & 0.98 & 0.98 & 0.98 & 0.98 \\ \cline{2-13} 
        ~ & \multirow{2}{*}{\textbf{ViT}} & AB & 0.93 & 0.86 & 0.90 & 0.89 & 0.90 & 0.92 & 0.91 & 0.93 & 0.92 & 0.95  \\ \cline{3-13}
        ~ & ~ & SGD & \textbf{0.99} & 0.95 & 0.98 & 0.96 & 0.97 & 0.95 & 0.95 & 0.98 & 0.98 & 0.97 \\ \hline \hline
         \multirow{4}{*}{\textbf{\rotatebox{90}{ImageNet}}} & \multirow{2}{*}{\textbf{VGG16}} & AB & \textbf{0.99} & \textbf{0.99} & \textbf{0.99} & \textbf{1.00} & \textbf{0.98} & \textbf{0.99} & \textbf{0.99} & \textbf{0.99} & \textbf{0.99} & \textbf{0.99} \\ \cline{3-13}
        ~ & ~ & SGD & 0.97 & 0.98 & 0.98 & 0.99 & 0.97 & 0.97 & 0.98 & 0.98 & 0.98 & \textbf{0.99} \\ \cline{2-13}
        ~ & \multirow{2}{*}{\textbf{ViT}} & AB & 0.90 & 0.86 & 0.90 & 0.86 & 0.89 & 0.92 & 0.90 & 0.93 & 0.91 & 0.94 \\ \cline{3-13}
        ~ & ~ & SGD & 0.98 & 0.94 & 0.97 & 0.95 & 0.96 & 0.96 & 0.96 & 0.98 & 0.97 & 0.98 \\ \hline
    \end{tabular}
    }
\end{table*}

\begin{table*}[t]
    \centering
    \caption{A detailed adversarial patch detection accuracy [0-1] of pure convolution (VGG) and attention (ViT) networks in conjunction with traditional classifiers in \textbf{unseen patch detection} setting.}
    \label{coco_ImageNet-unseen}
    \setkeys{Gin}{keepaspectratio}
\resizebox*{0.99\textwidth}{0.99\textheight} {
    \begin{tabular}{|l|l|l|l|c|c|c|c|c|c|c|c|c|c|}
    \hline
        \textbf{Dataset} & \textbf{Models} & \textbf{Classifier} & \textbf{Metric} & \textbf{Patch 0} & \textbf{Patch 1} & \textbf{Patch 2} & \textbf{Patch 3} & \textbf{Patch 4} & \textbf{Patch 5} & \textbf{Patch 6} & \textbf{Patch 7} & \textbf{Patch 8} & \textbf{Patch 9} \\ \hline \hline
       \multirow{8}{*}{\textbf{\rotatebox{90}{COCO}}} &  \multirow{4}{*}{\textbf{\rotatebox{90}{VGG16}}} & \multirow{2}{*}{LR} & Accuracy & 0.70 & 0.65 & 0.61 & 0.57 & 0.70 & 0.60 & 0.63 & 0.71 & 0.66 & 0.70 \\ \cline{4-14}
        ~ & ~ & ~ & STD & 0.10 & 0.13 & 0.16 & 0.12 & 0.09 & 0.14 & 0.13 & 0.16 & 0.14 & 0.17 \\ \cline{3-14}
        ~ & ~ & \multirow{2}{*}{SGD} & Accuracy & \textbf{0.80} & 0.78 & 0.64 & 0.59 & 0.79 & 0.67 & 0.71 & 0.75 & 0.67 & 0.72 \\ \cline{4-14}
        ~ & ~ & ~ & STD & 0.06 & 0.11 & 0.17 & 0.14 & 0.07 & 0.16 & 0.13 & 0.18 & 0.15 & 0.17 \\ \cline{2-14}
        ~ &  \multirow{4}{*}{\textbf{\rotatebox{90}{ViT}}} & \multirow{2}{*}{LR} & Accuracy & 0.67 & 0.81 & 0.74 & 0.73 & 0.83 & 0.72 & 0.71 & 0.72 & 0.72 & 0.68 \\ \cline{4-14}
        ~ & ~ & ~ & STD & 0.09 & 0.07 & 0.10 & 0.09 & 0.09 & 0.10 & 0.13 & 0.14 & 0.15 & 0.10 \\ \cline{3-14}
        ~ & ~ & \multirow{2}{*}{SGD} & Accuracy & 0.73 & \textbf{0.86} & \textbf{0.78} & \textbf{0.81} & \textbf{0.85} & \textbf{0.80} & \textbf{0.79} & \textbf{0.78} & \textbf{0.76} & \textbf{0.81} \\ \cline{4-14}
        ~ & ~ & ~ & STD & 0.08 & 0.05 & 0.09 & 0.07 & 0.08 & 0.09 & 0.12 & 0.12 & 0.13 & 0.08 \\ \hline \hline
        \multirow{8}{*}{\textbf{\rotatebox{90}{ImageNet}}} &  \multirow{4}{*}{\textbf{\rotatebox{90}{VGG16}}} & \multirow{2}{*}{LR} & Accuracy & 0.71 & 0.65 & 0.60 & 0.55 & 0.64 & 0.59 & 0.64 & 0.69 & 0.63 & 0.70 \\ \cline{4-14}
        ~ & ~ & ~ & STD & 0.12 & 0.15 & 0.15 & 0.12 & 0.08 & 0.13 & 0.16 & 0.17 & 0.13 & 0.17 \\ \cline{3-14}
        ~ & ~ & \multirow{2}{*}{SGD} & Accuracy & \textbf{0.73} & 0.74 & 0.63 & 0.57 & 0.73 & 0.65 & 0.74 & 0.73 & 0.68 & 0.73 \\ \cline{4-14}
        ~ & ~ & ~ & STD & 0.11 & 0.15 & 0.16 & 0.13 & 0.12 & 0.16 & 0.15 & 0.17 & 0.14 & 0.17 \\ \cline{2-14}
        ~ &  \multirow{4}{*}{\textbf{\rotatebox{90}{ViT}}} & \multirow{2}{*}{LR} & Accuracy & 0.64 & 0.78 & 0.71 & 0.77 & 0.83 & 0.69 & 0.71 & 0.68 & 0.69 & 0.65 \\ \cline{4-14}
        ~ & ~ & ~ & STD & 0.10 & 0.09 & 0.11 & 0.07 & 0.10 & 0.09 & 0.14 & 0.14 & 0.15 & 0.09 \\ \cline{3-14}
        ~ & ~ & \multirow{2}{*}{SGD} & Accuracy & \textbf{0.73} & \textbf{0.82} & \textbf{0.74} & \textbf{0.81} & \textbf{0.84} & \textbf{0.75} & \textbf{0.75} & \textbf{0.76} & \textbf{0.76} & \textbf{0.73} \\ \cline{4-14}
        ~ & ~ & ~ & STD & 0.10 & 0.07 & 0.10 & 0.07 & 0.08 & 0.10 & 0.12 & 0.13 & 0.12 & 0.10 \\ \hline
    \end{tabular}}
\end{table*}

\subsubsection{Without Noise}

The results, shown in Fig. (s) \ref{seenpatch} and \ref{unseenpatch}, provide a broad analysis of the average adversarial patch detection performance of each network under distinct patch conditions (seen and unseen). It is interesting to note that, under seen conditions, adversarial patches are almost perfectly detected when a pure convolutional neural network (without any form of attention layer) architecture (VGG16) is used. Out of all the classifiers used, the AdaBoost classifier yields the highest accuracy. The observation has been noted in both datasets, indicating that there is no bias in the adversarial patch detection performance. For example, when the AdaBoost (AB) classifier is attached to the features of VGG16, it yields $99.30$\% and $99.00$\% average classification performance COCO and ImageNet datasets, respectively. Random Forest (RF) yields the second-highest detection performance in seen-patch detection scenarios, with an accuracy gap of at most $0.30$\% compared to AdaBoost.

While it is expected that the detection performance of classifiers drops drastically in unseen (OOD) scenarios, surprisingly, the attention-based architecture (ViT) shows significant robustness against adversarial patches in OOD scenarios compared to the pure convolutional architecture (VGG). Further, the AB and RF classifiers, which are found highly effective in seen patch detection settings, exhibit the highest levels of vulnerability/reduction. Moreover, the SGD classifier is found to be highly robust at detecting adversarial patches in unseen patch-detection settings. For example, as shown in Fig. \ref{unseenpatch}, the SGD classifier achieves $79.90$\% and $76.90$\% average detection performance when ViT features are used to encode the real and attacked images, respectively. The SGD classifier, whether attached to VGG or ViT, yields the highest detection performance, making it agnostic to the feature extractor. Surprisingly, another simple classifier, namely logistic regression, shows greater robustness than other classifiers such as AdaBoost (AB), support vector machines (SVM), and random forests (RF). \textit{In brief, in terms of classifiers, AdaBoost (AB) and Random Forest (RF) are observed to be highly effective; whereas, in terms of the encoder, VGG16 outperforms ViT. However, in the setting of unseen patches, ViT outperforms VGG, SGD, and Logistic Regression (LR) and is found highly resilient as compared to other classifiers.} It is to note here that we have also evaluated the effectiveness of other deep encoders such as NASNetMobile \cite{zoph2018learning} and Xception \cite{chollet2017xception}, but found VGG outperforms them by a significant margin. For example, the performance of VGG on COCO and ImageNet is at least $5.4\%$ better than that of NASNetMobile. Similarly, other traditional classifiers, such as k-nearest neighbour (KNN), decision trees, and gradient boosting, are evaluated but found less effective than AB and SGD; hence, results using only the best classifiers are reported in this paper.

Table \ref{coco_ImageNet-seen} demonstrates the detailed performance of each encoder using two best-performing classifiers, namely AB and SGD. As mentioned above, in the seen-patch detection setting, perfect patch-detection accuracy is observed when the VGG encoder is used with the AB classifier. When the ViT encoder is used with the SGD classifier, at least 94\% accuracy is observed across the dataset, demonstrating that it is easy to detect patches when they are seen at test time. However, \textit{`we believe such high accuracy can provide a false sense of security'} because it is hard to predict the future set of adversarial patches, and hence detection algorithm must be effective under those \textbf{zero-shot patches}. Henceforth, to demonstrate how easy (challenging) it is to detect adversarial patches, \textit{we have performed $10$ fold unseen patch cross-validation experiments}. In this setting, in every fold, images of a single patch along with clean images are used for training, and images of the other nine patches along with clean images are used for testing. The detailed performance of these experiments in terms of average accuracy along with standard deviation (STD) is reported in Table \ref{coco_ImageNet-unseen}.

First and foremost, it is observed that the classifiers that are showing perfect detection performance in the seen setting suffer a drastic reduction in performance in the zero-shot (unseen) patch setting. Secondly, the SGD classifier outperforms the AB classifier in the unseen-patch robustness scenario, and ViT surpasses the VGG encoder's performance. It is also worth noting that ViT, along with SGD, not only yields high average accuracy but also shows lower standard deviation. Among all the patches, patch$_1$ and patch$_4$ (Fig. \ref{fig:dataset}) are found to be highly effective at detecting unseen patches and achieve at least $85$\% and $82$\% average accuracy on the COCO and ImageNet datasets, respectively. 

\begin{figure}[t]
\centering
  \includegraphics[width = 0.49\textwidth]{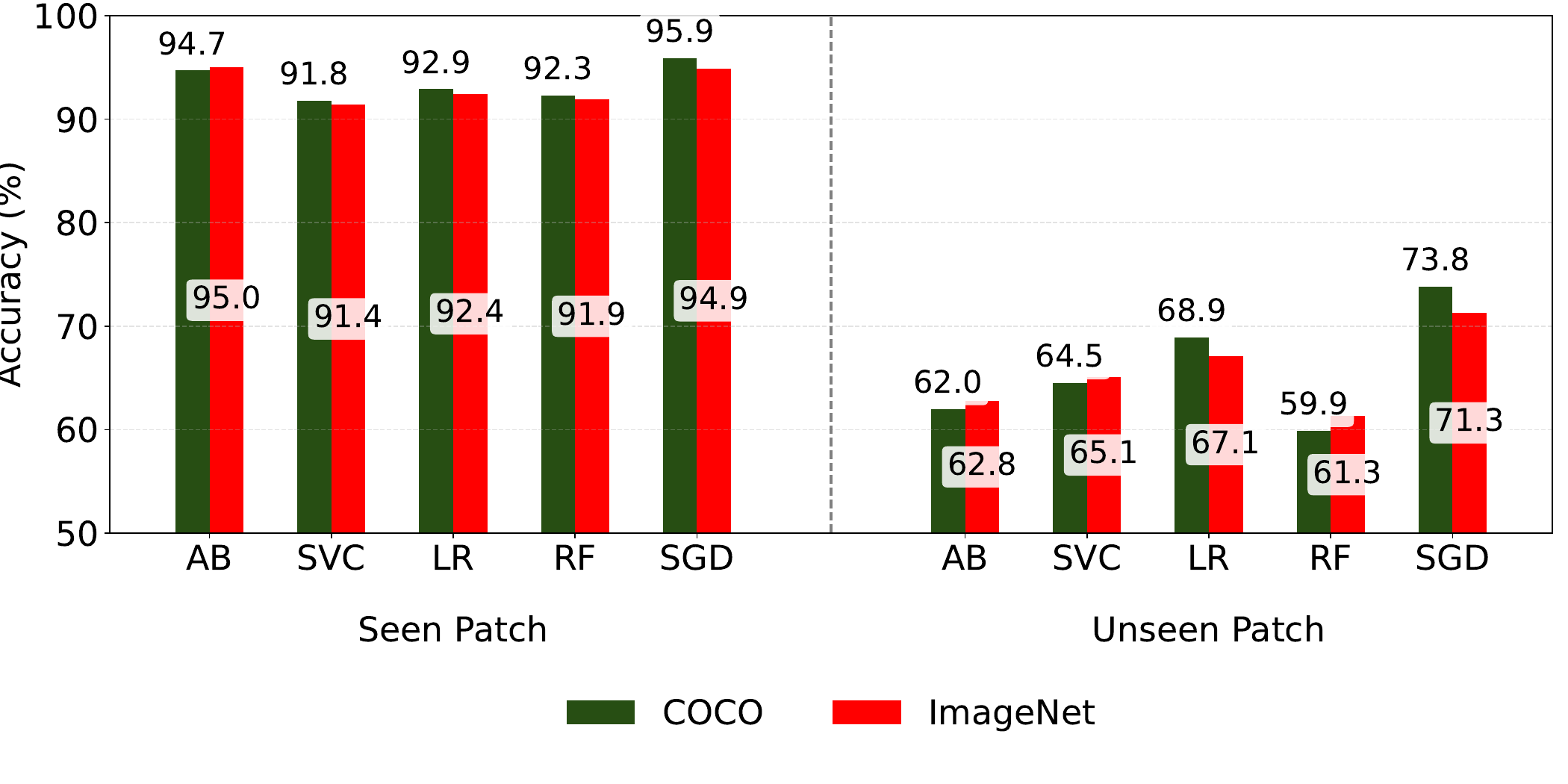}
  \caption{Average adversarial patch attack detection in the presence of noise corruptions. Performances are reported on VGG16 in the seen patch setting (left) and ViT in the unseen patch setting  (right).}
    \label{fig:GSInoise}
\end{figure}

\begin{table*}[!ht]
    \centering
\caption{Adversarial patch detection accuracy [0-1] on COCO and ImageNet in seen patch but unseen noise evaluation setting. In other words, the resiliency of the detectors is evaluated when a seen patch perturbed with natural noises comes for classification.}
    \label{vgg16coco_ImageNet-seen}
    \setkeys{Gin}{keepaspectratio}
\resizebox*{0.99\textwidth}{0.99\textheight} {
    \begin{tabular}{|l|l|l|l|c|c|c|c|c|c|c|c|c|}
    \hline
        \textbf{Dataset} & \textbf{Models} & \textbf{Classifier} & \textbf{Patch 0} & \textbf{Patch 1} & \textbf{Patch 2} & \textbf{Patch 3} & \textbf{Patch 4} & \textbf{Patch 5} & \textbf{Patch 6} & \textbf{Patch 7} & \textbf{Patch 8} & \textbf{Patch 9} \\ \hline \hline
        
        \multirow{2}{*}{\textbf{COCO}} & \multirow{2}{*}{\textbf{VGG16}} & AB & 0.92 & \textbf{0.99} & \textbf{0.98} & \textbf{0.98} & 0.89 & 0.94 & 0.97 & \textbf{0.95} & 0.89 & 0.95 \\ \cline{3-13}
        ~ & ~ & SGD & \textbf{0.94} & 0.98 & 0.96 & \textbf{0.98} & \textbf{0.91} & \textbf{0.97} & \textbf{0.98} & \textbf{0.95} & \textbf{0.96} & \textbf{0.97} \\ \hline \hline
        \multirow{2}{*}{\textbf{ImageNet}} & \multirow{2}{*}{\textbf{VGG16}} & AB & \textbf{0.95} & \textbf{0.98} & \textbf{0.98} & \textbf{0.98} & \textbf{0.88} & 0.95 & \textbf{0.98} & 0.96 & \textbf{0.92} & 0.93 \\ \cline{3-13}
        ~ & ~ & SGD & \textbf{0.95} & \textbf{0.98} & 0.97 & 0.97 & 0.83 & \textbf{0.97} & \textbf{0.98} & \textbf{0.97} & 0.91 & \textbf{0.98} \\ \hline
    \end{tabular}
    }
\end{table*}

\begin{table*}[t]
    \centering
    \caption{Adversarial patch detection accuracy [0-1] along with standard deviation (STD) on COCO and ImageNet in an unseen patch and unseen noise evaluation setting. In other words, the `dual' resiliency of the detectors is evaluated when unseen patches perturbed with natural noises come for classification.}
    \label{NASMobileNetcoco_ImageNet-unseen}
    \setkeys{Gin}{keepaspectratio}
\resizebox*{0.99\textwidth}{0.99\textheight} {
    \begin{tabular}{|l|l|l|l|c|c|c|c|c|c|c|c|c|c|}
    \hline
        \textbf{Dataset} & \textbf{Models} & \textbf{Classifier} & \textbf{Metric} & \textbf{Patch 0} & \textbf{Patch 1} & \textbf{Patch 2} & \textbf{Patch 3} & \textbf{Patch 4} & \textbf{Patch 5} & \textbf{Patch 6} & \textbf{Patch 7} & \textbf{Patch 8} & \textbf{Patch 9} \\ \hline \hline
        \multirow{4}{*}{\textbf{\rotatebox{90}{COCO}}} & \multirow{4}{*}{\textbf{\rotatebox{90}{ViT}}} & LR & Accuracy & 0.58 & 0.74 & 0.66 & 0.67 & \textbf{0.80} & 0.72 & 0.66 & 0.67 & 0.74 & 0.65  \\ \cline{4-14}
        ~ & ~ & ~ & STD & 0.08 & 0.08 & 0.10 & 0.09 & 0.08 & 0.10 & 0.12 & 0.11 & 0.12 & 0.09 \\ \cline{3-14}
        ~ & ~ & SGD & Accuracy & \textbf{0.60} & \textbf{0.77} & \textbf{0.71} & \textbf{0.75} & \textbf{0.80} & \textbf{0.75} & \textbf{0.73} & \textbf{0.73} & \textbf{0.74} & \textbf{0.79} \\ \cline{4-14}
        ~ & ~ & ~ & STD & 0.08 & 0.05 & 0.10 & 0.07 & 0.06 & 0.08 & 0.11 & 0.12 & 0.11 & 0.08 \\ \hline \hline
        \multirow{4}{*}{\textbf{\rotatebox{90}{ImageNet}}} & \multirow{4}{*}{\textbf{\rotatebox{90}{ViT}}} & LR & Accuracy & 0.59 & 0.71 & 0.64 & 0.70 & \textbf{0.80} & 0.66 & 0.65 & 0.68 & 0.68 & 0.61  \\ \cline{4-14}
        ~ & ~ & ~ & STD & 0.08 & 0.09 & 0.10 & 0.07 & 0.08 & 0.09 & 0.12 & 0.12 & 0.14 & 0.07\\ \cline{3-14}
        ~ & ~ & SGD & Accuracy & \textbf{0.66} & \textbf{0.73} & \textbf{0.66} & \textbf{0.76} & 0.78 & \textbf{0.70} & \textbf{0.68} & \textbf{0.73} & \textbf{0.75} & \textbf{0.68} \\ \cline{4-14}
        ~ & ~ & ~ & STD & 0.09 & 0.09 & 0.10 & 0.06 & 0.08 & 0.10 & 0.12 & 0.11 & 0.11 & 0.09 \\ \hline
    \end{tabular}
    }
\end{table*}

\subsubsection{Resiliency Under Noise Perturbation}

In this setting, we have evaluated the resiliency of adversarial patch detectors trained on clean patches (which did not see any form of noise during training). The prime reason is that natural noises are inherent in the environment \cite{Agarwal_2020_CVPR_Workshops} and, like patches, every form of noise cannot be used for training; hence, trained detectors must be resilient to handle unseen natural noises. As observed above, VGG and ViT are found to be effective in seen and unseen patches, respectively; only their resiliency is evaluated in the respective settings. The results of the resiliency of VGG and ViT in seen and unseen patch detection settings are reported in Fig. \ref{fig:GSInoise}. While it is expected that the detectors will suffer drops in detection performance, surprisingly, only a marginal reduction is observed. For example, the performance of SGD drops from $97.8$\% to $95.9$\% and from $97.9$\% to $94.9$\% on the COCO and ImageNet datasets, respectively. Interestingly, in the seen patch evaluation setting under noise, the SGD classifier outperforms the best classifier, i.e., AB found in the seen patch without the noise setting. Similar to seen patch settings, a reduction of $5-6$\% in the accuracy of SGD is observed in unseen noise patch settings from unseen without noise patch settings. The detailed performance in the seen path noise setting is reported in Table \ref{vgg16coco_ImageNet-seen}, which shows that on the COCO dataset, SGD yields the best performance most of the time; whereas on the ImageNet AB, SGD performs comparably, with an average accuracy difference of $0.1$\%. 

Table \ref{NASMobileNetcoco_ImageNet-unseen} showcases the average $10$ fold cross-validation performance in the unseen patch noise evaluation setting. Interestingly, when the images (clean and patched) are perturbed and the evaluation has been performed in a zero-shot setting, logistic regression (LR) outperforms the AdaBoost (AB) classifier. However, the SGD classifier plugged into the features of ViT outperforms each classifier and encoder by a significant margin. Overall, through an extensive experimental evaluation, it is observed that the \textit{VGG encoder along with the AB classifier yields the highest effectiveness} in detecting adversarial patches but the constrained is these patches must be seen during detector training. Moreover, the \textit{ViT encoder along with the SGD classifier is not only found generalized in unseen patch evaluation settings but also yields high resiliency} when the images are perturbed through noise corruptions. Therefore, \textbf{we assert that in real-world settings, a defender should pick the ViT encoder along with the SGD classifier to defend against adversarial patches}. It is to be noted since the evaluation of traditional classifiers and such detailed generalized settings are missing in the literature, the proposed research can pave a path for future research aiming to defend against adversarial patches and noises.

\subsection{Comparison with Baseline}

\begin{figure}[t]
\centering
  \includegraphics[width = 0.49\textwidth]{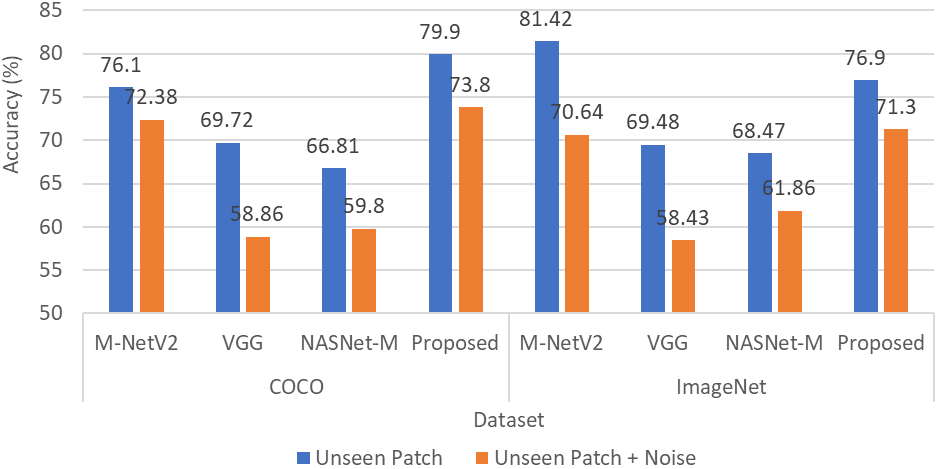}
  \caption{Comparison with the SOTA in terms of average adversarial patch detection accuracy for unseen patch and unseen patch + unseen noise detection.}
    \label{fig:baseline}
     \vspace{-7mm}
\end{figure}

As mentioned above, in the literature, limited work has been done so far on adversarial patch detection containing a variety of patches \cite{ojaswee2023benchmarking}. To demonstrate the strength of the proposed work, we have now performed the comparison with recent state-of-the-art (SOTA) adversarial patch detection work by Ojaswee et al. \cite{ojaswee2023benchmarking}. For comparison, we have fine-tuned the models reported as best in their paper, such as MobileNet-V2 (M-NetV2) and NASNet-MobileNet (NASNet-M), and compared them with the proposed generalized and robust detection network (ViT + SGD). As showcased in Fig. \ref{fig:baseline}, the proposed algorithm outperforms each SOTA except M-NetV2 on the ImageNet. However, it is observed that the standard deviation (STD)  of the proposed algorithm is significantly lower ($3.46$\%) than that of the best-performing existing model on ImageNet.

\section{WHY ROBUST PATCH DETECTION MATTERS?}

While, we have comprehensively demonstrated the transferability of adversarial patches across classification networks, one might argue that, since classification models are trained with similar loss functions, they might be inherently vulnerable even if they are not seen during patch generation. Therefore, we explicitly study the transferability of patch+noise attacks against unseen tasks and architectures, i.e., object detection. Furthermore, we demonstrate that robustness inheritance in patch detectors is difficult to achieve with standard data augmentation or noise injection strategies.

\subsection{Transferability of Patch+Noise Attacks to Object Detection}
To further examine the transferability of the proposed patch–noise attack, we extend our evaluation beyond image classification to the object detection task. All experiments in this setting are conducted on the COCO validation set. Specifically, we assess whether adversarial patches combined with natural noise corruptions remain effective when transferred to object detection architectures. As shown in Table~\ref{tab:detection_patch_noise}, the combined patch–noise attack consistently degrades object detection performance across multiple state-of-the-art object detectors, even though the patches are not optimised for these models or the detection task. Compared to clean inputs, patch-only attacks already reduce mAP, and adding noise leads to a larger drop in all cases. For instance, Faster R-CNN with a ResNet-50 backbone drops from 36.88 mAP on clean images to 33.62 under patch-only attacks, and further to 20.89 when patch and noise are combined. Similar trends are observed for YOLO-based detectors, where YOLO-v8 decreases from 36.69 to 15.87 mAP and YOLO-v11 from 45.66 to 25.54 mAP under patch–noise attacks. These results indicate that the proposed patch–noise combination transfers across unseen architectures and from image classification to object detection.

\begin{table}[t]
\centering
\caption{Vulnerability of state-of-the-art object detection models to adversarial patch attacks and their combination with natural noise. Results are reported as mAP (higher is better) on the COCO validation set.}
\label{tab:detection_patch_noise}
\setkeys{Gin}{keepaspectratio}
\resizebox*{0.49\textwidth}{!}{
{\small
\setlength{\tabcolsep}{6pt}
\renewcommand{\arraystretch}{1.2}
\begin{tabular}{|l|c|c|c|}
\hline
\multicolumn{1}{|c|}{\textbf{Model}} &
\textbf{Clean} &
\textbf{Patch} &
\textbf{Patch + Noise} \\
\hline \hline
Faster R-CNN (ResNet-50) & 36.88 & 33.62 & 20.89 \\ \hline
YOLO-v8                 & 36.69 & 33.41 & 15.87 \\ \hline
YOLO-v9                 & 51.87 & 47.49 & 34.70 \\ \hline
YOLO-v10                & 38.22 & 35.20 & 19.35 \\ \hline
YOLO-v11                & 45.66 & 41.51 & 25.54 \\
\hline
\end{tabular}
}}
\end{table}

\subsection{Effect of Data Augmentation and Noise Injection on Patch Detection}

We further analyze whether standard data augmentation and noise injection strategies can account for the observed robustness and generalization gains. We train our best-performing configuration, using a ViT backbone with an SGD-trained classifier, and evaluate it under generalized settings involving unseen adversarial patches and unseen patch–noise combinations. As reported in Table~\ref{tab:augmentation_ablation}, commonly used augmentation techniques such as RandAugment, Cutout, and strong colour–geometry transformations achieve comparable performance across both evaluation settings. However, none of these approaches consistently outperforms our method across the two metrics. For instance, RandAugment achieves 0.86\% accuracy under unseen patch evaluation and 0.78\% under unseen patch + noise, while Cutout attains 0.83\% and 0.81\%, respectively. Our method achieves 0.85\% and 0.80\%, indicating competitive performance without relying on complex augmentations.

In addition, we study the effect of injecting random noise during training, a common strategy for improving robustness. Using the same ViT-based configuration trained with random noise on the best-performing patch (patch 4), we evaluate on unseen patches (patch IDs 0–9, excluding 4) and unseen noise types. Under this setting, performance decreases from 85.26\% to 79.81\%, indicating that random noise injection during training does not improve robustness to unseen adversarial patches and noise. These results suggest that the primary gains in our framework stem from the proposed singularity-aware training objective rather than from data augmentation or noise injection alone.

\begin{table}[t]
\centering
\caption{Impact of data augmentation on generalization to unseen adversarial patches and robustness under unseen patch + noise settings.}
\label{tab:augmentation_ablation}
\setkeys{Gin}{keepaspectratio}
\resizebox*{0.49\textwidth}{!}{
{\small
\setlength{\tabcolsep}{6pt}
\renewcommand{\arraystretch}{1.2}
\begin{tabular}{|l|c|c|}
\hline
\multicolumn{1}{|c|}{\textbf{Augmentation}} &
\textbf{Unseen Patch} &
\textbf{Unseen Patch + Noise} \\
\hline \hline
RandAugment            & 0.86 & 0.78 \\ \hline
Cutout                 & 0.83 & 0.81 \\ \hline
Strong Color + Geometry & 0.86 & 0.81 \\ \hline
Ours                   & 0.85 & 0.80 \\
\hline
\end{tabular}
}}
\end{table}

\section{CONCLUSION}

Adversarial patches are among the strongest forms of adversaries in the physical world, and are agnostic to multiple transformations such as rotation and translation. On a similar note, natural noises such as Gaussian and Impulse noise are inherently present in images due to the unconstrained environment. Due to these stealthy attacks, the development of deep neural networks in the physical world is risky and lacks trustworthiness. Interestingly, the defense against adversarial attacks failed to adequately address adversarial patches. Further, no research addresses patches and noise simultaneously, leaving the existence of a unified defense in jeopardy. Henceforth, in this research, we not only present a benchmark adversarial patch and noise-perturbed dataset but also present a detailed benchmark study. The study reveals several interesting observations and provides a classifier that is an effective defense against adversarial patches and natural noise. In the future, we aim to expand the dataset by incorporating additional manipulations, including adversarial perturbations, to facilitate the development of a universal defence mechanism. Along with the dataset, a novel attention-guided self-supervised patch-detection algorithm will be presented as a unified defense solution. 

\subsection*{Acknowledgements}

V. Kumar is partially supported through the Visvesvaraya PhD Fellowship.

\subsection*{Dataset Release}

The dataset will be released at \url{https://tbvl22.github.io/website/resources.html}

\small{
\bibliographystyle{apalike}
\bibliography{aistats}
}

\section*{Checklist}

\begin{enumerate}

 \item For all models and algorithms presented, check if you include:
 \begin{enumerate}
   \item A clear description of the mathematical setting, assumptions, algorithm, and/or model. [Yes/No/Not Applicable]   \textbf{Yes}
   \item An analysis of the properties and complexity (time, space, sample size) of any algorithm. [Yes/No/Not Applicable]  \textbf{Yes}
   \item (Optional) Anonymized source code, with specification of all dependencies, including external libraries. [Yes/No/Not Applicable]
 \end{enumerate}

 \item For any theoretical claim, check if you include:
 \begin{enumerate}
   \item Statements of the full set of assumptions of all theoretical results. [Yes/No/Not Applicable] \textbf{Not Applicable}
   \item Complete proofs of all theoretical results. [Yes/No/Not Applicable]  \textbf{Not Applicable}
   \item Clear explanations of any assumptions. [Yes/No/Not Applicable]  \textbf{Not Applicable}   
 \end{enumerate}

 \item For all figures and tables that present empirical results, check if you include:
 \begin{enumerate}
   \item The code, data, and instructions needed to reproduce the main experimental results (either in the supplemental material or as a URL). [Yes/No/Not Applicable]  \textbf{Yes}
   \item All the training details (e.g., data splits, hyperparameters, how they were chosen). [Yes/No/Not Applicable]  \textbf{Yes}
    \item A clear definition of the specific measure or statistics and error bars (e.g., with respect to the random seed after running experiments multiple times). [Yes/No/Not Applicable]  \textbf{Yes}
    \item A description of the computing infrastructure used. (e.g., type of GPUs, internal cluster, or cloud provider). [Yes/No/Not Applicable]  \textbf{Yes}
 \end{enumerate}

 \item If you are using existing assets (e.g., code, data, models) or curating/releasing new assets, check if you include:
 \begin{enumerate}
   \item Citations of the creator If your work uses existing assets. [Yes/No/Not Applicable]  \textbf{Yes}
   \item The license information of the assets, if applicable. [Yes/No/Not Applicable]  \textbf{Not Applicable}
   \item New assets either in the supplemental material or as a URL, if applicable. [Yes/No/Not Applicable]  \textbf{Not Applicable}
   \item Information about consent from data providers/curators. [Yes/No/Not Applicable]  \textbf{Not Applicable}
   \item Discussion of sensible content if applicable, e.g., personally identifiable information or offensive content. [Yes/No/Not Applicable]  \textbf{Not Applicable}
 \end{enumerate}

 \item If you used crowdsourcing or conducted research with human subjects, check if you include:
 \begin{enumerate}
   \item The full text of instructions given to participants and screenshots. [Yes/No/Not Applicable]  \textbf{Not Applicable}
   \item Descriptions of potential participant risks, with links to Institutional Review Board (IRB) approvals if applicable. [Yes/No/Not Applicable]  \textbf{Not Applicable}
   \item The estimated hourly wage paid to participants and the total amount spent on participant compensation. [Yes/No/Not Applicable]  \textbf{Not Applicable}
 \end{enumerate}

 \end{enumerate}

\end{document}